\documentclass{article}

\usepackage{iclr2027_conference,times}

\usepackage{amsmath,amsfonts,bm}

\def\eqref#1{equation~\ref{#1}}

\def\1{\bm{1}}

\DeclareMathAlphabet{\mathsfit}{\encodingdefault}{\sfdefault}{m}{sl}
\SetMathAlphabet{\mathsfit}{bold}{\encodingdefault}{\sfdefault}{bx}{n}

\usepackage{graphicx}
\usepackage{wrapfig}
\usepackage{booktabs}
\usepackage{makecell}
\usepackage{pifont}
\usepackage{multirow}
\usepackage{float}
\usepackage[table]{xcolor}
\usepackage{hyperref}
\hypersetup{hidelinks}

\usepackage{capt-of}
\usepackage{url}

\newcommand{\cmark}{\textcolor{green!60!black}{\ding{51}}}
\newcommand{\xmark}{\textcolor{red}{\ding{55}}}
\definecolor{mywarm1}{RGB}{255,248,220}
\definecolor{mywarm2}{RGB}{255,235,205}
\definecolor{mywarm3}{RGB}{255,228,225}
\definecolor{deepgreen}{RGB}{0,70,0}

\title{Tac-DINO: Learning Tactile Features with Patch Alignment}
\author{Hong Li\textsuperscript{1}, Yankang Dong\textsuperscript{1}, Yue Xu\textsuperscript{1}, Yihan Tang\textsuperscript{1}, Jiamin Qiu\textsuperscript{1}, Qihang Yao\textsuperscript{1}, \\ \textbf{Xing Zhu\textsuperscript{2}, Yujun Shen\textsuperscript{2}, Nan Xue\textsuperscript{2}, Yong-Lu Li\textsuperscript{1, 3}}\thanks{Correspondence to: Yong-Lu
Li $<$yonglu\_li@sjtu.edu.cn$>$.} \\
\textsuperscript{1}Shanghai Jiao Tong University, \textsuperscript{2}Ant Group, \textsuperscript{3}Shanghai Innovation Institute\\
\texttt{\selectfont\{hong\_li, yonglu\_li\}@sjtu.edu.cn}
}

\iclrfinalcopy

\begin{document}

\maketitle
\fancyhead{}
\renewcommand{\headrulewidth}{0pt}

\begin{abstract}
    Multisensory perception is essential for robotic manipulation, with tactile sensing providing critical information for fine-grained, contact-rich interactions. Although prior work integrates tactile and spatial signals to improve perception, vision-tactile learning commonly relies on object-level alignment. Addressing this limitation requires systematically aligned 3D vision-tactile data. We develop a data collection system and introduce \textbf{Touch3D}, a dataset comprising \textbf{505 real-world objects} and \textbf{20,025 tactile contacts}. Building on this dataset, we establish the \textbf{Contact-Centric Multimodal Tactile Representation Benchmark (CoMT-Bench)}, which expands evaluation from object-level retrieval and tactile properties to include patch-level contact evaluation. We further propose \textbf{Geometry-Relation Distilled Tactile-Centric Cross-Modal Learning (GeoTAC)}, which explores tactile learning across all multisensory data. 
    Experiments on the Touch3D test set, patch-level alignment improves the average score by \textbf{5.32} over object-level alignment, with GeoTAC's geometric supervision improving \textbf{4.48} more. In robotic manipulation, GeoTAC improves TP by \textbf{19.14} and \textbf{11.34} over DINOv2 (Vision-only) and DINOv2 (Vision-Tactile), respectively.
\par\noindent\textbf{Project page:} \url{https://lihongcs.github.io/tacdino}.
\end{abstract}

\section{Introduction}
\label{sec:intro}

Human perceive the world through multiple senses, including vision, hearing, and touch~\citep{hedger2025vicarious}. Inspired by this, multi-directional breakthroughs have emerged in various artificial intelligence applications. Computer vision now excels at object understanding, driven by the emergence of numerous 2D image datasets~\citep{lin2014microsoft, li2023beyond, li2024isolated}. 3D vision has overcome data bottlenecks with the rise of Objaverse~\citep{deitke2023objaverse} and Objaverse-XL~\citep{deitke2024objaverse}. Recently, tactile sensing become a more critical modality especially for robotics. The development of diverse sensors~\citep{taylor2022gelslim,bhirangi2024anyskin} and hardware systems has yielded substantial amounts of paired vision-tactile data.
Here different modalities share intrinsic physical properties, such as geometric shape, material composition, and texture. Therefore, it is vital to connecting these corresponding modalities for a joint understanding of objects and environment, particularly for robotics, virtual reality and next generation physical intelligence.

\begin{figure}[!t]
\centering
\includegraphics[width=0.80\linewidth]{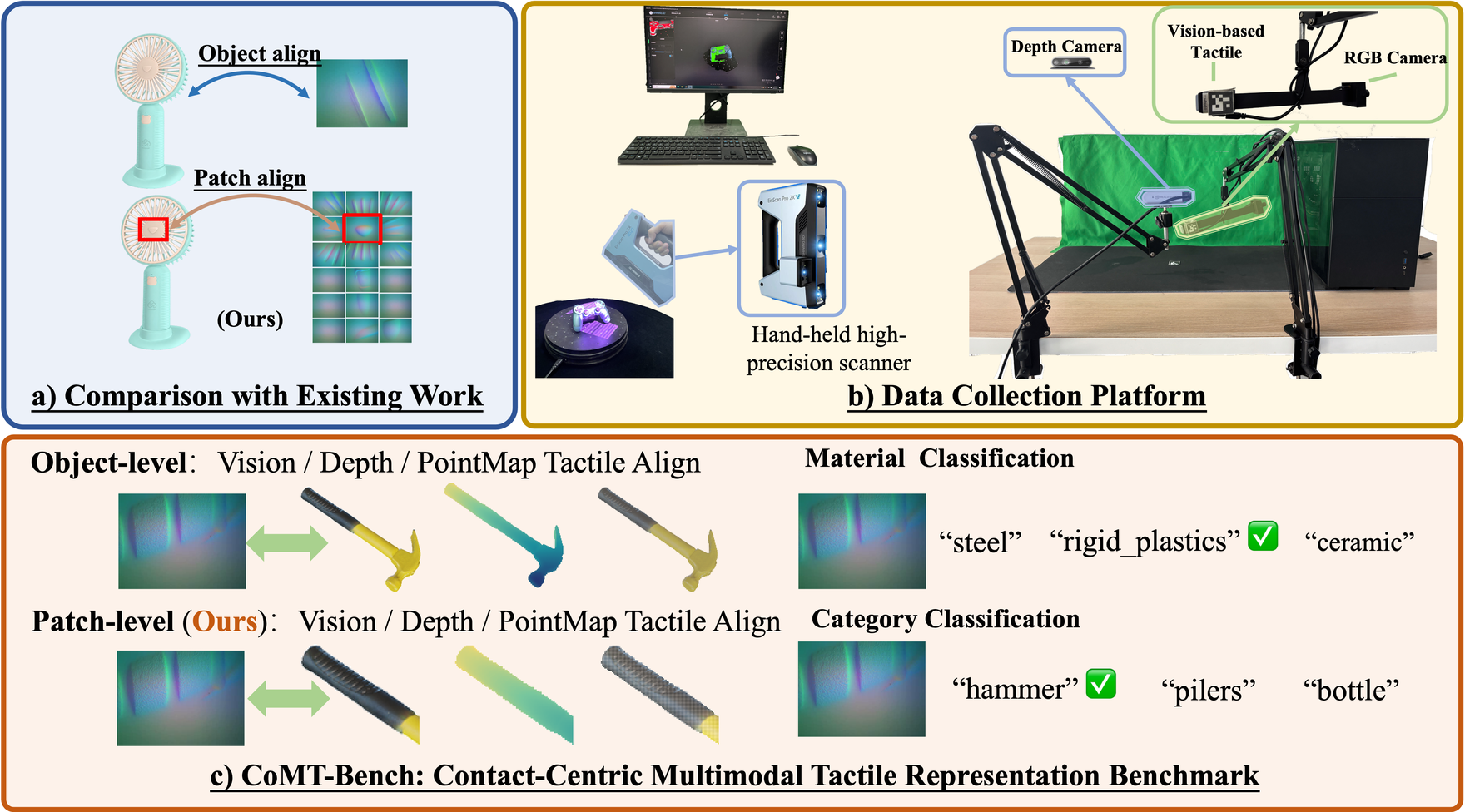}
\caption{Overview of the Tac-DINO framework. (a) We focus on patch-level alignment rather than the object-level alignment used in existing work. (b) We develop a data collection platform for 3D vision-tactile data. (c) Building on existing work, we propose CoMT-Bench to evaluate tactile representations using multisensory data.}
\label{fig:teaser}
\end{figure}

Multimodal large language models have achieved substantial success
in question-answering tasks~\citep{liu2023visual}, with similar
progress in vision-language-action (VLA) models for generalizable manipulation~\citep{kim2024openvla}. Despite these advances, limited physical and spatial understanding can still lead to hallucinations and interaction inconsistencies~\citep{li2023evaluating}. Existing work incorporates tactile~\citep{li2019connecting,huang2025tactile} and 3D representations~\citep{wang2024rise} to enhance contact-rich manipulation and spatial perception. The efficiency of incorporating these modalities suggests that next-generation models will continue to incorporate more modalities toward human-like intelligence. However, current research lacks both a formal definition of the relationships between tactile sensing and other modalities and a unified approach that integrates all relevant modalities for tactile learning. The main difficulty is the lack of large-scale 3D vision-tactile datasets collected under a formally defined and systematic protocol. ObjectFolder Real closely matches these requirements, but its limited scale of 100 objects~\citep{gao2023objectfolder} constrains broader methodological exploration.

To this end, we first develop a data collection system and collect a 3D vision-tactile dataset, \textbf{Touch3D}, comprising \textbf{505 objects} and \textbf{20,025 tactile samples}, together with scanned object geometry, contact-location annotations, and material labels (Fig.~\ref{fig:teaser}(b)). We then introduce the \textbf{Contact-Centric Multimodal Tactile Representation Benchmark (CoMT-Bench)} (Fig.~\ref{fig:teaser}(c)), which formally associates each tactile observation with a physically localized region in RGB, depth, and pointmap data, rather than an object image with an undefined scale and contact correspondence  (Fig.~\ref{fig:teaser}(a)). CoMT-Bench evaluates both object-level and patch-level alignment between tactile and other perceptual modalities, as well as tactile-based material and object-category classification. Together, these evaluations provide a common basis for comparing existing methods and studying tactile-centric representation learning.

Furthermore, to investigate tactile representation learning with patch-aligned multi-sensory data, we propose \textbf{Geometry-Relation Distilled Tactile-Centric Cross-Modal Learning (GeoTAC)}. GeoTAC combines DINO local-to-global self-distillation, tactile-centric cross-modal alignment, and geometry distillation that transfers the geometric relationships between tactile observations and object surfaces. Through extensive experiments, we investigate three questions: (1) Does patch-level alignment improve tactile performance compared with object-level? (2) Can depth and point clouds, when incorporated through appropriate methods improve tactile representation learning? (3) Do the learned tactile representations consistently improve performance in downstream robotic manipulation tasks? Our experiments lead to the following conclusions:

Our main contributions are threefold:
\textbf{(1)} We formally define the spatial correspondence between tactile observations and visual and geometric modalities, and develop a data-collection system for constructing a large-scale 3D vision-tactile dataset.
\textbf{(2)} We introduce CoMT-Bench, which evaluates tactile representations through object-level alignment, patch-level multimodal alignment, and tactile property recognition, including material and object-category classification.
\textbf{(3)} We propose GeoTAC, a tactile-centric cross-modal representation learning method. GeoTAC improves over object-level alignment by 5.32, gains a further 4.48 by incorporating depth and pointmaps. In robotics mainpulation task, it improves average TP by 19.14 over DINOv2 (Vision-only), and by 11.34 over DINOv2 (Vision-Tactile), respectively.

\section{Related work}

\subsection{Multi-source Tactile data}
Tactile sensors provide contact cues essential for dexterous manipulation. Representative technologies include visuo-tactile~\citep{gelsight,2020DIGIT,2024gelsim4.0,ren2023mc}, capacitive~\citep{glauser2019deformation,wu2020capacitivo,xu2016stretch}, resistive~\citep{sundaram2019learning,bhattacharjee2013tactile,stassi2014flexible}, and magnetic~\citep{bhirangi2025anyskin} sensing. Visuo-tactile sensors are particularly prevalent because they are inexpensive and easy to fabricate. ObjectFolder~\citep{gao2021objectfolder,gao2022objectfolder,gao2023objectfolder} provides visual and acoustic signals, X-Capture~\citep{clarke2025xcapture} includes material annotations, and Touch100K~\citep{cheng2024touch100k} introduces multi-granularity language labels.

\subsection{Vision-Tactile Learning}
Recent vision-tactile learning tackles representation integration and sensor heterogeneity across three main directions. The first is learning composite representations through imitation learning~\citep{heng2025vitacformer, romero2024eyesight, lin2024generalize, pattabiraman2024learning}, which updates parameters via action alignment. The second transfers tactile perception into intermediate representations, for example, by converting it into point clouds~\citep{huang20253dvitac, yuan2024robot} or reconstructing it with NeRF~\citep{suresh2024neuralfeels, dou2024tactile}. The last is learning tactile features using self-supervised methods~\citep{feng2025anytouch, feng2026anytouch, rodriguez2025contrastive,guzey2023dexterity,cheng2026taco, higuera2026visuo}. Despite these advances, tactile signals are typically captured with only coarse alignment to vision, neglecting the precise alignment between the tactile data and the contact patch area. In this work, we propose to study vision-tactile patch alignment feature learning.

\subsection{Representation Learning}

Robust representation learning from multimodal data is critical for robotic perception, driving the adoption of self-supervised learning (SSL) paradigms like DINO~\citep{dino1st}, CLIP~\citep{radford2021clip}, V-JEPA~\citep{v-jepa1}, MAE~\citep{he2021mae}, and GAN~\citep{goodfellow2014gan}. Among these, DINO distinctly excels in discriminative feature extraction. Unlike methods focused on reconstruction, generation, or cross-modal alignment, DINO's self-distillation mechanism captures both fine-grained local details and holistic global semantics. DINOv2~\citep{oquab2023dinov2} further scales this framework to massive corpora. Since the local-to-global paradigm perfectly aligns with the requirement of capturing local tactile contacts within a global scene context, we utilize DINOv2 as our base model to investigate vision-tactile patch alignment learning.

\section{Touch3D Dataset}

\begin{figure}
    \centering
    \includegraphics[width=0.9\linewidth]{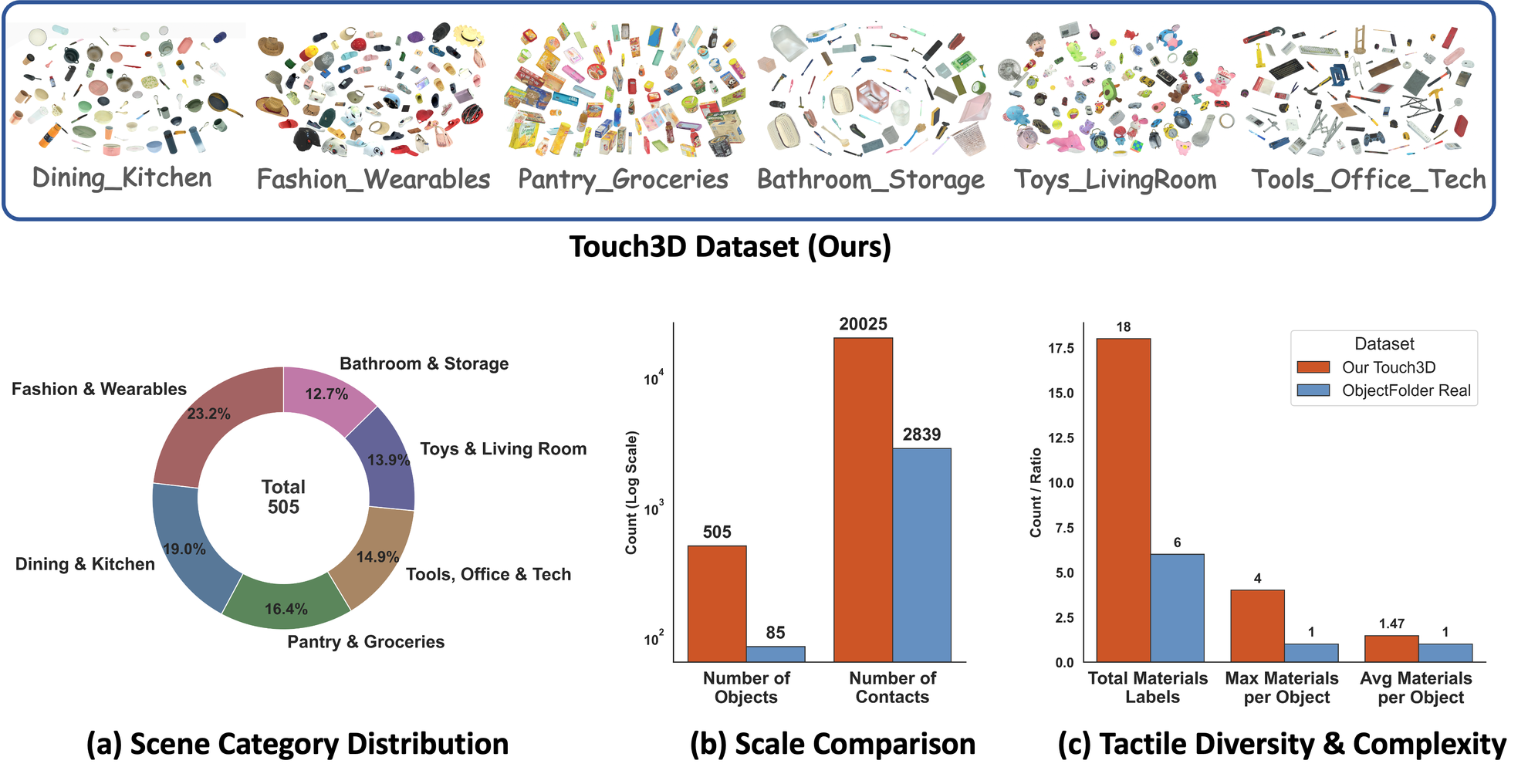}
    \caption{Data distribution and labeling comparison.}
    \label{fig:data_distribution.}
\end{figure}

Multisensory learning is critical for representation learning in fine-grained robotic manipulation. Tactile, depth, and pointmap observations are commonly integrated with vision to address the limitations of vision-only systems in contact-rich interaction and spatial perception. However, collecting real-world multisensory data is expensive and time-consuming.
ObjectFolder Real~\citep{gao2023objectfolder} collects data from 100 real-world objects~\footnote{Data obtained from \url{https://objectfolder.stanford.edu/}; after filtering out incompletely labeled data, 85 objects with 2,839 contacts remained usable.} to mitigate this challenge. However, its limited object diversity restricts its suitability for large-scale representation learning, making it more appropriate for evaluation than training. The dataset was collected using a Franka Emika Panda robot arm equipped with a GelSight visuo-tactile sensor~\citep{dong2017improved,yuan2017gelsight}. This setup requires carefully controlled physical interactions and is difficult to scale because data collection relies on time-consuming teleoperation. Therefore, to address data scarcity and support multisensor-enhanced tactile learning, we introduce a movable, time-efficient data collection system capable of handling arbitrary real-world objects. We then propose \textbf{Touch3D}, a dataset comprising \textbf{505} real objects and \textbf{20,025} tactile contacts.

\subsection{3D Shape Data Collection}
\label{subsec: 3d scanning}
We use an EinScan Pro 2X 2020 handheld 3D scanner to collect a high-quality 3D mesh that includes point locations and corresponding color textures for each object. The scanner utilizes structured light scanning, which involves three steps to construct high-quality 3D features: projecting grating patterns, capturing the deformed grating, and calculating the 3D point cloud.  The minimum point distance of the produced 3D point cloud reaches 0.2 mm, enabling a high-definition and precise reconstruction of the physical form. For each object, we scan over two faces, some complex objects scan 4 $\sim$ 5 faces, and then combine them into one 3D shape. After merging, we utilize the official scanning software to filter out noise points, as shown in Fig.~\ref{fig:teaser} b).

\subsection{Vision-Tactile Data Collection}
\label{subsec: image-tactile}
We utilize the GelSight Mini as our tactile sensor to construct a vision-tactile data collection platform. The platform consists of an omnidirectional stand mounted on a desktop, a custom 3D-printed holder attached to the stand, and a GelSight Mini sensor alongside an RGB camera fixed at opposite ends of this holder. Additionally, a depth camera is mounted on a separate omnidirectional stand to record the collection process, providing third-view data.
For each object, we first use the tactile sensor to record the pressing process, accompanied by a third-person perspective video captured by the depth camera. Next, we rotate the 3D-printed sensor holder by 180 degrees while keeping all other components stationary. Once a paired data sample is collected, the data collection system displays the scanned 3D shape, and the operator manually selects the contact location. After completing these three steps to gather a set of paired image-tactile data, we move to the next contact location to continue the collection process.

\begin{figure}[t]
\centering
\includegraphics[width=1.0\linewidth]{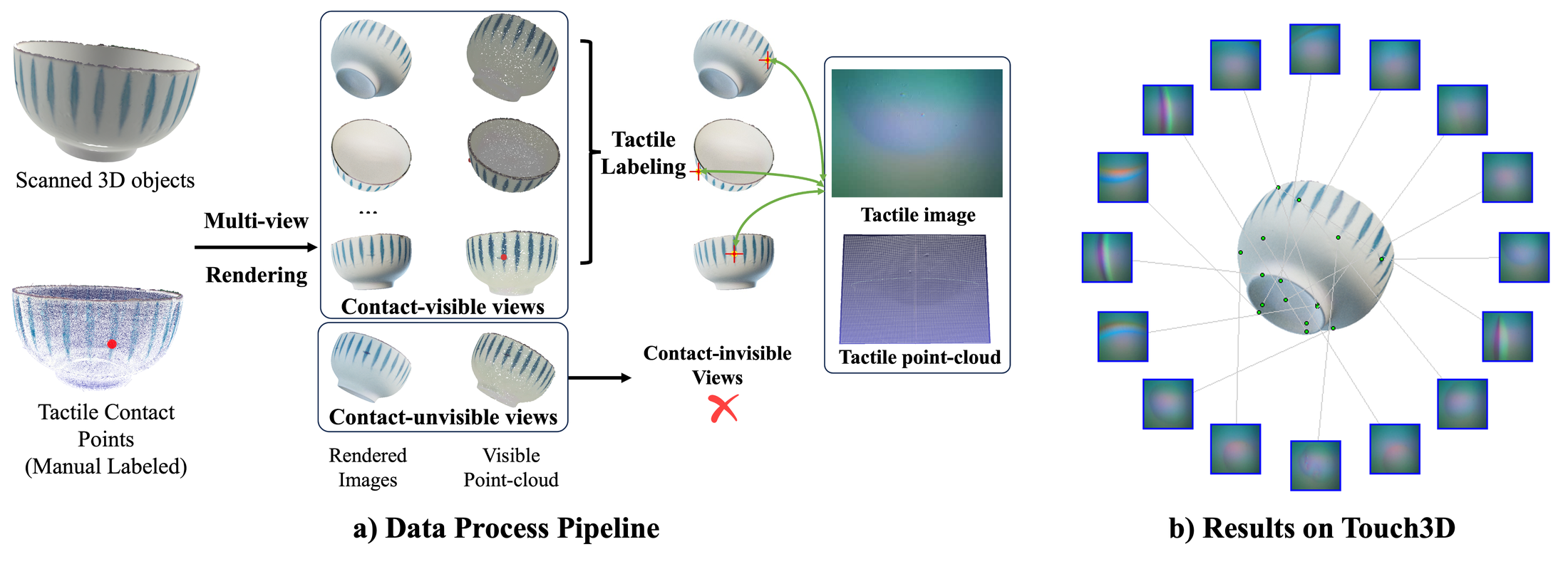}
\caption{\textbf{Contact-region data curation.} Scanned shapes, registered views, and contact-center annotations associate a tactile contact with a local visual region. }
\label{fig:data_process}
\end{figure}

\subsection{Human Annotation and Sanity Check}
\label{sec: human annotation}

To label the contact locations and tactile materials, as well as to double-check the quality of the scanned meshes, tactile, and vision data, we developed a data labeling system.
The system includes features to load captured data, add or select tactile material labels, and show the contact location annotations. When the scanned 3D shape includes noisy points or the contact location exhibits bias, we utilize MeshLab~\citep{cignoni2011meshlab} to filter out the noise or correct the labels.  When vision or tactile data is damaged or missing, we filter it out and then renumber the remaining data.

\subsection{Data Statistics}

\begin{table}[!t]
\caption{Comparison of multisensory datasets in terms of collection environments, numbers of objects and tactile contacts, and included modalities.}
\label{tab:data_comparison}
\begin{center}
\setlength{\tabcolsep}{2pt}
\resizebox{0.95\linewidth}{!}{%
\begin{tabular}{@{}lcccccc@{}}
\toprule
\multirow{2}{*}{Dataset} & \multirow{2}{*}{\textbf{Env}} & \multirow{2}{*}{\textbf{Obj.}} & \multirow{2}{*}{\textbf{Tac.}} & \multirow{2}{*}{\makecell{Tactile\\Sensor}} & \multicolumn{2}{c}{Correlated Modalities} \\
\cmidrule{6-7}
& & & & & RGB-D & Scanned 3D \\
\midrule
Touch and Go~\citep{yang2022touch} & Real & 3,971 & 13.9k & Gelsight & \xmark & \xmark \\
Objectfolder 2.0~\citep{gao2022objectfolder} & Simulation & 1,000 & - & Gelsight & \cmark & \cmark \\
X-Capture~\citep{clarke2025x} & Real & 500 & 3,000 & DIGIT & \cmark & \xmark \\
ObjectFolder Real~\citep{gao2023objectfolder} & Real & 100 & 2,995 & Gelsight & \cmark & \cmark \\
\midrule
\textbf{Touch3D (Ours)} & Real & \textbf{505} & \textbf{20,025} & \textbf{Gelsight Mini} & \cmark & \cmark \\
\bottomrule
\end{tabular}%
}
\end{center}
\end{table}

Using our data collection and labeling system, we referenced the object selection from PACE~\citep{you2024pace} and collected \textbf{505} objects with \textbf{20,025} tactile contacts. The overall time cost was over \textbf{700} hours, including \textbf{200} hours to scan 3D shapes at a rate of 2-3 objects per hour, and \textbf{500} hours to collect vision-tactile data.To the best of our knowledge, our \textbf{Touch3D} dataset, is the largest 3D Vision-Tactile dataset in the community, as shown in Table~\ref{tab:data_comparison}.  A comparison with ObjectFolder Real~\citep{gao2023objectfolder} can be found in Fig.~\ref{fig:data_distribution.}. The ObjectFolder Real features kitchen objects. Our dataset includes 6 different scenarios: \textit{dining\_kitchen, fashion\_wearables, pantry\_groceries, bathroom\_storage, toys\_livingroom, and tools\_office\_tech}. Each scenario has a balanced number of objects. After eliminating the visual indivisibility tactile material label, our dataset includes 18 materials, compared to 6 in ObjectFolder Real. In ObjectFolder Real, each object has only one material. Our dataset includes more diverse objects, with each object having at most 4 materials, a mean of 1.47 material labels.

\section{Methods}

Multi-sensory learning enhances perceptual understanding, while integrating tactile sensing with depth and point maps is critical for contact-rich manipulation and spatial understanding in robotics. The effectiveness of joint multimodal learning encourages next-generation models to incorporate an increasing number of modalities. Motivated by these observations, we first formally define the relationship between tactile sensing and other modalities. We then introduce the Contact-Centric Multimodal Tactile Representation Benchmark (\textbf{CoMT-Bench}) (Sec.~\ref{subsec: vis-tac benchmark}), which extends tactile evaluation from object-level alignment to patch-level alignment. Finally, to investigate tactile-centric representation learning from patch-aligned multisensory data, we propose Geometry-Relation Distilled Tactile-Centric Cross-Modal Learning (\textbf{GeoTAC}) (Sec.~\ref{subsec: patch alignment methods}).

\subsection{CoMT-Bench: Contact-Centric Multimodal Tactile Representation Benchmark}
\label{subsec: vis-tac benchmark}

Existing tactile learning benchmarks commonly evaluate object-level cross-modal alignment and material or category classification using manually annotated labels, without a standardized definition of the visual region corresponding to each tactile observation. We extend this evaluation from object-level to patch-level alignment to comprehensively evaluate tactile features.

For each tactile contact at $\mathbf{p}$, we extract aligned observations from RGB, depth, and pointmap inputs: $X_{\mathrm{local}}^{m} = \operatorname{Crop} ( X^{m},\pi_m(\mathbf{p}) ),\quad m\in\{V,D,P\}$, where $\pi_m$ projects the contact location into modality $m$. Given a local tactile query $T_{\mathrm{local}}$, object-level evaluation retrieves a global observation from modality $m$: $G_m^{*} = \operatorname*{arg\,max}_{G\in\mathcal{G}_m} \operatorname{Sim} ( E_T(T_{\mathrm{local}}), E_m(G) )$, where $\mathcal{G}_m$ is the global candidates. Object-level retrieval is evaluated using object-category kNN. Patch-level evaluation replaces $\mathcal{G}_m$ with the local candidate set and treats only the observation from the same contact as positive. We additionally evaluate material prediction and object-category classification using the tactile representation alone.

\subsection{\textbf{GeoTAC}: Geometry-Relation Distilled Tactile-Centric Cross-Modal Learning}
\label{subsec: patch alignment methods}

\begin{figure}[t]
\centering
\includegraphics[width=0.8\linewidth]{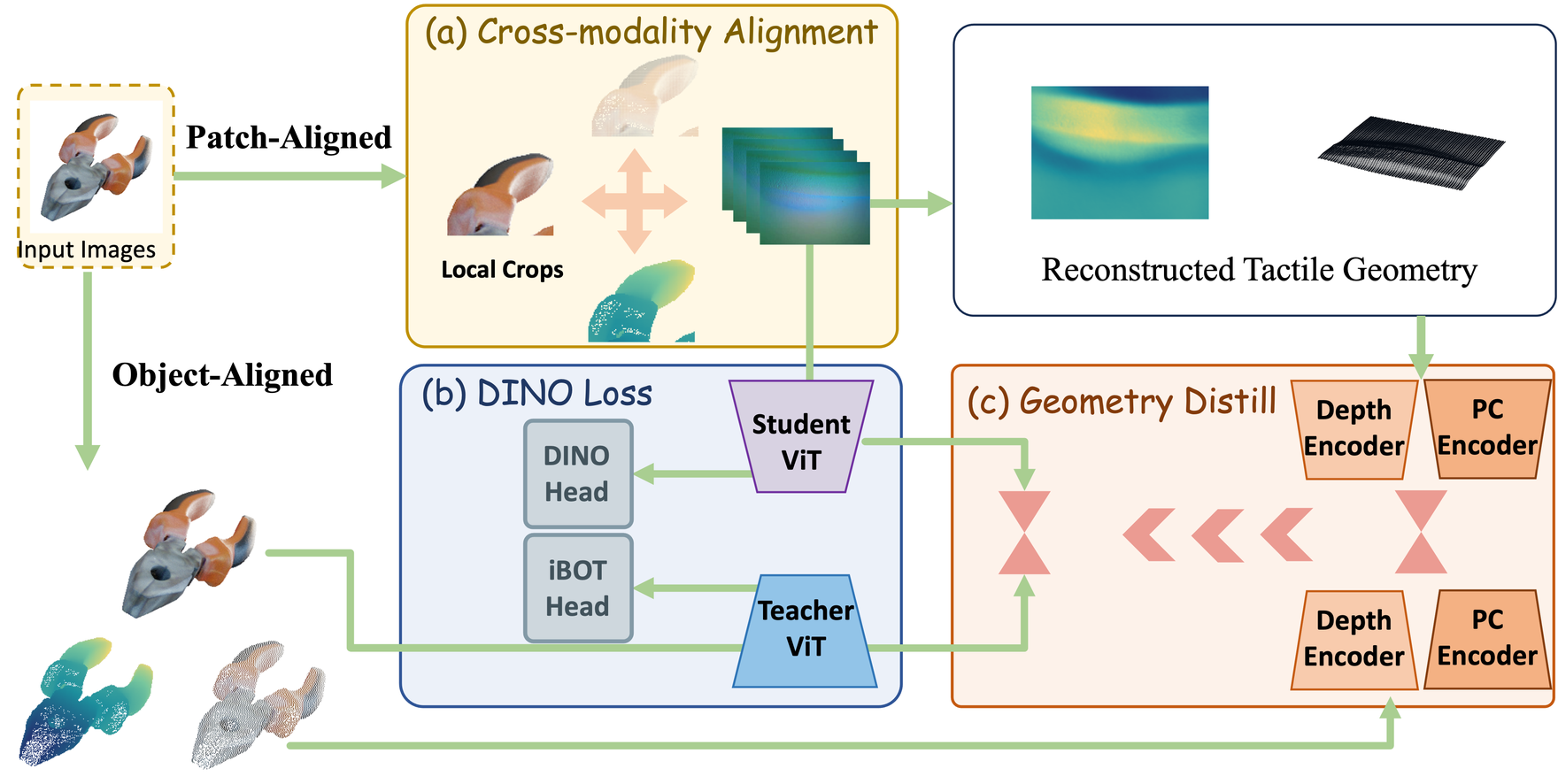}
\caption{Overview of the proposed GeoTAC framework.}
\label{fig:method}
\end{figure}

We learn tactile representations from patch-aligned RGB, depth, and pointmap observations. As shown in Figure~\ref{fig:method}, GeoTAC consists of tactile-centered cross-modal alignment, the retained DINOv2 objective, and geometry distillation.

\paragraph{Tactile-Centered Cross-Modal Alignment.}
We align local RGB, depth, and pointmap observations with the tactile representation of the same contact: $\mathcal{L}_{\mathrm{align}} = \frac{1}{3} [ \mathcal{L}_{c}(V_l,T_l) + \mathcal{L}_{c}(D_l,T_l) + \mathcal{L}_{c}(P_l,T_l) ]$, where $\mathcal{L}_{c}$ is the bidirectional contact-level contrastive loss. It uses soft region matching rather than comparing patches with identical indices. The external RGB, depth, and point encoders are frozen, while the tactile encoder, temporal head, and projection heads are optimized.

\paragraph{Local-to-Global Distillation.}
Following DINOv2~\citep{oquab2023dinov2}, the teacher processes global RGB crops, while the student processes the corresponding local tactile observations:
\begin{equation}
\mathcal{L}_{\mathrm{lg}}
=
-\frac{1}{N_gN_l}
\sum_{g=1}^{N_g}
\sum_{l=1}^{N_l}
P_{\mathrm{teacher}}(V_g)^{\mathsf{T}}
\log P_{\mathrm{student}}(T_l),
\label{eq:local_global_distillation}
\end{equation}
where $P_{\mathrm{teacher}}$ and $P_{\mathrm{student}}$ are the DINO-head output distributions. The original global-crop DINO, iBOT, and KoLeo objectives are retained, giving $\mathcal{L}_{\mathrm{DINO}} = \mathcal{L}_{\mathrm{lg}} + \mathcal{L}_{\mathrm{gg}} + \mathcal{L}_{\mathrm{iBOT}} + \mathcal{L}_{\mathrm{KoLeo}}$, where $\mathcal{L}_{\mathrm{gg}}$ denotes the global RGB DINO loss.

\paragraph{Geometry-Relation Distillation.}
We reconstruct tactile depth $\widetilde{D}_l$ and tactile point cloud $\widetilde{P}_l$ and compare them with the corresponding local depth $D_l$ and pointmap $P_l$. Their equally weighted bidirectional similarity distributions form the geometry teacher: $R_{\mathrm{geo}}^{r} = \frac{1}{2} ( R_{D,\widetilde{D}}^{r} + R_{P,\widetilde{P}}^{r} ),\quad {r\in\{\rightarrow,\leftarrow\}}$. The geometry relations are distilled into the local RGB--tactile matching distribution $R_{V,T}^{r}$:
\begin{equation}
\mathcal{L}_{\mathrm{geo}}
=
\frac{1}{2}
\sum_{r\in\{\rightarrow,\leftarrow\}}
D_{\mathrm{KL}}
\left(
\left[
\operatorname{softmax}
\left(
R_{\mathrm{geo}}^{r}/\tau
\right)
\right]
\,\middle\|\,
\operatorname{softmax}
\left(
R_{V,T}^{r}/\tau
\right)
\right).
\label{eq:geometry_distillation}
\end{equation}
The reconstructed tactile depth and point cloud are used only during training.

The complete GeoTAC objective is $\mathcal{L}_{\mathrm{GeoTAC}} = \mathcal{L}_{\mathrm{DINO}} + \lambda_{\mathrm{align}} \mathcal{L}_{\mathrm{align}} + \lambda_{\mathrm{geo}} \mathcal{L}_{\mathrm{geo}}$, where $\lambda_{\mathrm{align}}$ and $\lambda_{\mathrm{geo}}$ control the contributions of cross-modality alignment and geometry distillation.

\section{Experiments and Discussion}

\subsection{Experimental Settings}

\begin{wrapfigure}{r}{0.55\linewidth}
    \centering
    \includegraphics[width=\linewidth]{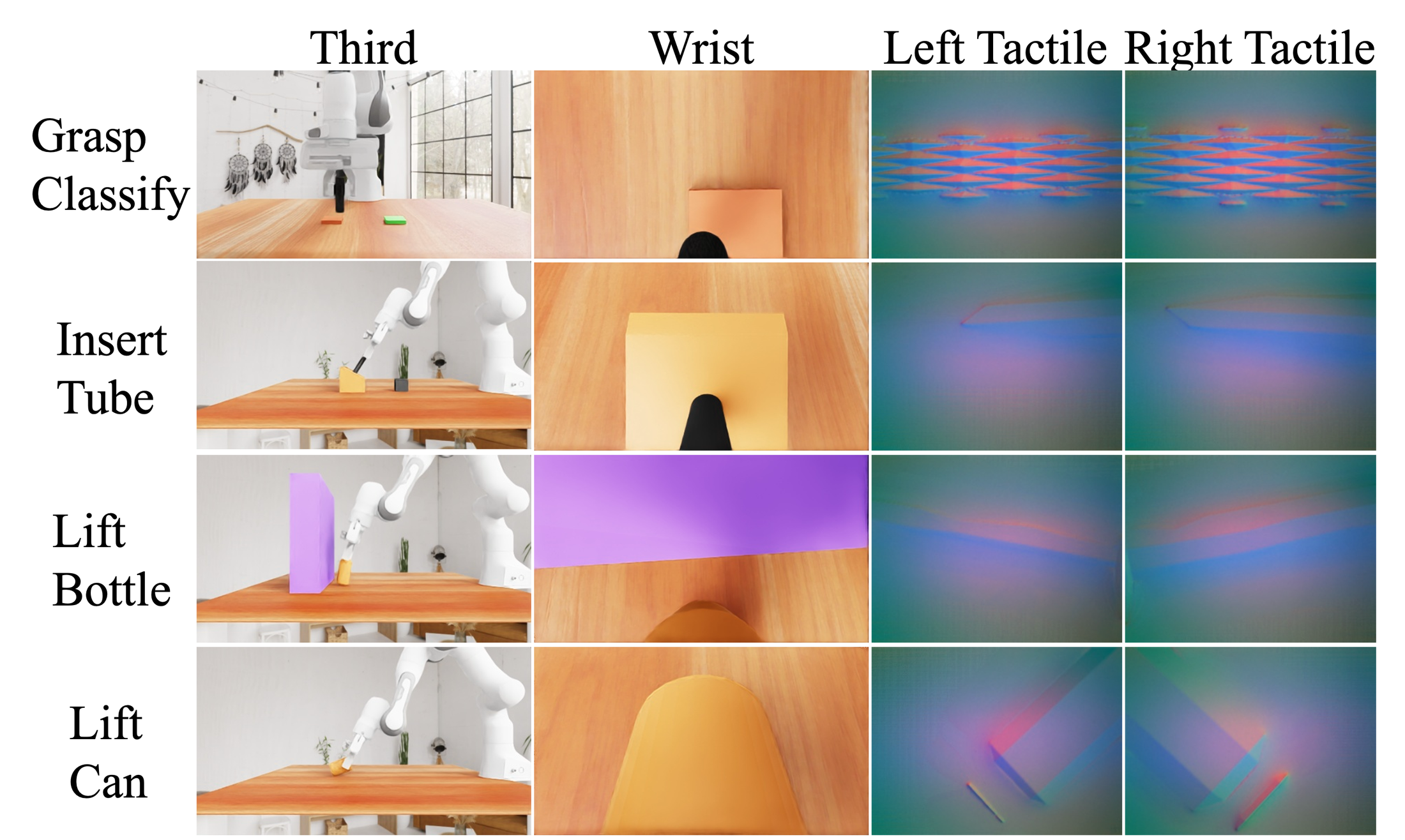}
    \caption{Visualization of the UniVTAC evaluation.}
    \label{fig:univtac}
\end{wrapfigure}

We systematically evaluate patch alignment using DINOv2~\citep{oquab2023dinov2} ViT-Large models. The collected data are split into training, validation, and test sets at 4:1:1. We also use ObjectFolder Real as an additional test set. We include general vision-tactile methods, Contrastive Learning (CL)~\citep{radford2021learning} and DAR~\citep{aytar2017see}, as baselines. We further compare multisensory alignment methods, including ImageBind~\citep{girdhar2023imagebind}, UniTouch~\citep{yang2024binding}, and ViT-Lens~\citep{lei2024vitlens}, for tactile learning using all available multisensory data. For modalities without an encoder in these methods, we incorporate general-purpose encoders, such as PointBERT.

We also use the UniVTAC~\citep{chen2026univtac} simulator as a testbed for evaluating robotic manipulation, with the GelSight Mini tactile sensor matching that used in Touch3D. Following the official configuration, we select four tasks, as shown in Figure~\ref{fig:univtac}, and collect 100 demonstration episodes per task to train Diffusion Policy~\cite{chi2025diffusion} for 30,000 steps.

\subsection{Metrics}
\label{subsec: metrics}

\paragraph{Tactile Representation Evaluation.}
Object-level retrieval is measured by object-category k-NN Top-5 accuracy. Patch-level alignment is measured by exact-contact Top-5 retrieval accuracy (E@5), averaged over both retrieval directions. For tactile properties, we report k-NN mAP for multi-label material classification and probe accuracy for object-category classification. \textit{Avg.} denotes the arithmetic mean of the reported metrics.

\paragraph{Robotic Manipulation.}
We initialize each policy with the pretrained encoders and run 40 rollouts using the official UniVTAC settings. We report success rate (SR) and average final task progress (TP, \%). \textit{Avg.} is the mean across the four tasks, computed before rounding.

\subsection{Effectiveness of Patch-Aligned Tactile Learning}

In this subsection, we investigate whether patch-level alignment is more effective than object-level alignment for tactile representation learning. We focus on the vision-tactile setting, comparing object-level vision-tactile alignment, patch-aligned tactile learning, and patch-aligned tactile learning combined with vision-tactile methods.

Table~\ref{tab:vision-tactile} summarizes the results of vision-tactile learning methods. From the results, we can draw the following conclusions:
(1) Vision-tactile alignment is helpful for tactile feature learning (see all results compared with the Tac row).
(2) Patch alignment improves performance compared with object-level alignment. The Tac + CL row shows lower patch-level performance than the Tac-only baseline. Patch Tac + CL achieves the best results.
(3) The results on ObjectFolder Real are consistent with those on Touch3D, but overall performance is lower due to cross-sensor generalization.

\begin{table}[t]
\caption{Tactile representation learning with object-, patch-level, and tactile-property evaluation.}
\label{tab:vision-tactile}
\begin{center}
\setlength{\tabcolsep}{2pt}
\begin{tabular}{@{}l*{4}{c}|c@{}}
\toprule
\multirow{3}{*}{Training configuration}
& Object-level
& Patch-level
& \multicolumn{2}{c|}{Tactile properties}
& \multirow{3}{*}{Avg. $\uparrow$} \\
\cmidrule(lr){2-2}
\cmidrule(lr){3-3}
\cmidrule(lr){4-5}
& $T_l \to V_g$
& $T_l \leftrightarrow V_l$
& Material
& Category
& \\
& Top-5
& E@5
& mAP
& Acc.
& \\
\midrule
\multicolumn{6}{l}{\textbf{Touch3D}} \\
Tac
& 27.49 & 4.32 & 25.94 & 56.73
& 28.62 (0.00) \\
Tac + CL~\citep{radford2021learning}
& 37.68 & 3.72 & 28.81 & 58.98
& 32.30 \textcolor{green!50!black}{($+3.68$)} \\
Patch Tac
& 46.50 & 6.46 & 30.37 & 56.42
& 34.94 \textcolor{green!50!black}{($+6.32$)} \\
Patch Tac + CL~\citep{radford2021learning}
& 42.94 & 13.95 & 34.50 & 59.08
& 37.62 \textcolor{green!50!black}{($+9.00$)} \\
Patch Tac + DAR~\citep{aytar2017see}
& 47.39 & 7.62 & 31.72 & 56.81
& 35.89 \textcolor{green!50!black}{($+7.27$)} \\
\midrule
\multicolumn{6}{l}{\textbf{ObjectFolder Real}} \\
Tac
& 26.41 & 1.92 & 14.97 & 0.07
& 10.84 (0.00) \\
Tac + CL~\citep{radford2021learning}
& 27.13 & 1.82 & 15.95 & 0.33
& 11.31 \textcolor{green!50!black}{($+0.47$)} \\
Patch Tac
& 30.26 & 1.92 & 16.62 & 1.51
& 12.58 \textcolor{green!50!black}{($+1.74$)} \\
Patch Tac + CL~\citep{radford2021learning}
& 30.77 & 3.85 & 21.71 & 2.13
& 14.62 \textcolor{green!50!black}{($+3.77$)} \\
Patch Tac + DAR~\citep{aytar2017see}
& 33.08 & 2.05 & 17.85 & 3.12
& 14.03 \textcolor{green!50!black}{($+3.18$)} \\
\bottomrule
\end{tabular}
\end{center}
\end{table}

\subsection{Effectiveness of Tactile-Centric Cross-Modal Learning}

In this subsection, we investigate two questions: (1) Does incorporating additional modalities improve performance over vision-tactile learning? (2) Is our proposed tactile-centric learning method more effective than treating all modalities equally during alignment?

\begin{figure}[!t]
    \centering
    \includegraphics[width=1.0\linewidth]{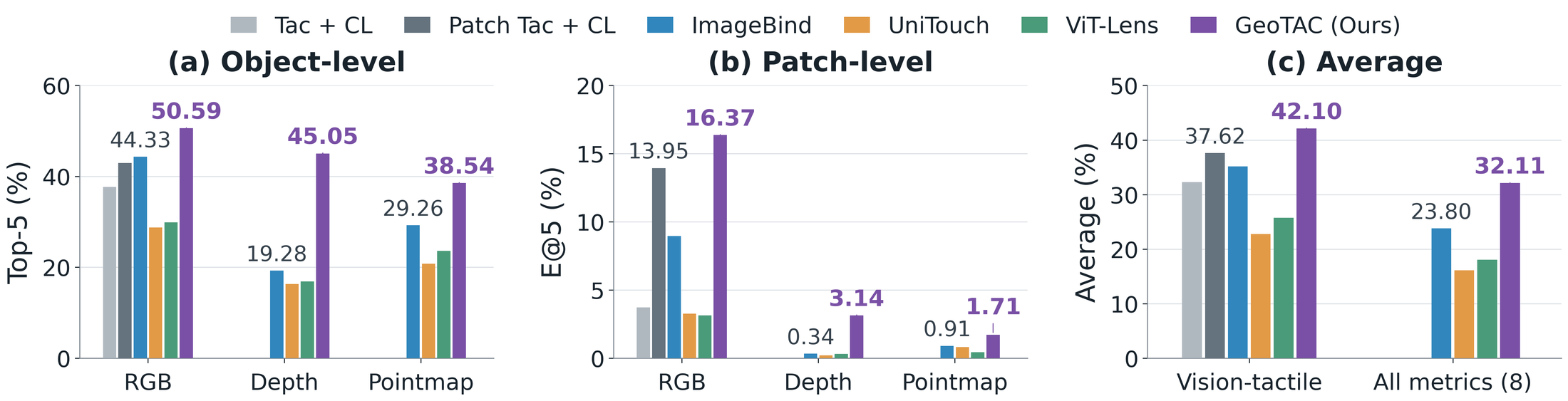}
    \caption{Visualization of multisensory learning results on CoMT-Bench.}
    \label{fig:multi-sensory-result}
\end{figure}

\begin{figure}[t]
\centering
\begin{minipage}[t]{0.62\linewidth}
    \vspace{0pt}
    \centering
    \includegraphics[width=0.85\linewidth]{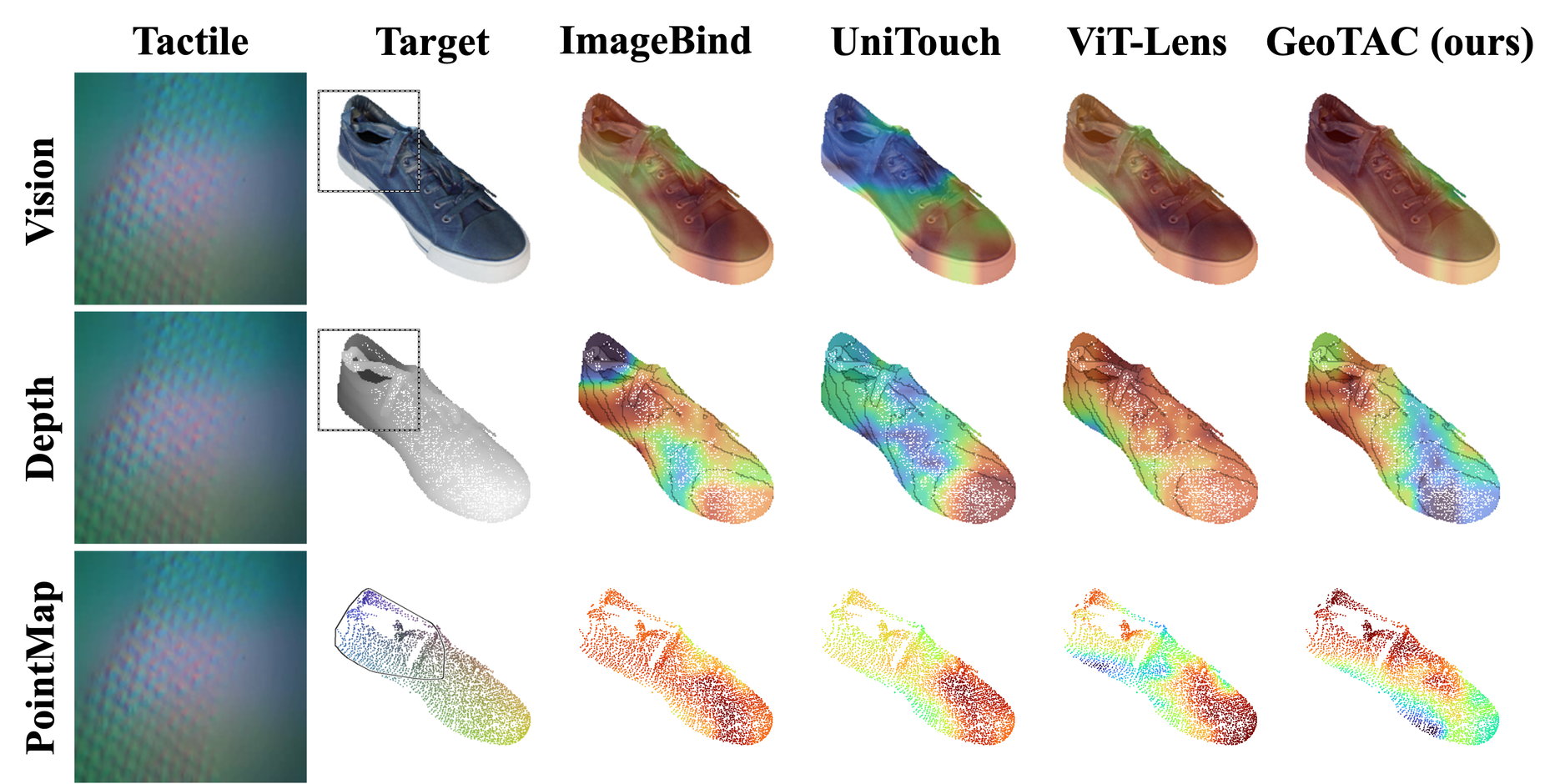}
    \caption{Attention visualization for a Touch3D sample across multisensory learning methods.}
    \label{fig:attention}
\end{minipage}
\hfill
\begin{minipage}[t]{0.35\linewidth}
    \vspace{0pt}
    \centering
    \captionof{table}{Ablation study of GeoTAC on Touch3D.}
    \label{tab:ablation-study}
    \begin{center}
\setlength{\tabcolsep}{4pt}
    \renewcommand{\arraystretch}{1.0}
        \begin{tabular}{@{}lc@{}}
    \toprule
    Training configuration & Avg. $\uparrow$ \\
    \midrule
    Patch Tac       & 34.94 \\
    Patch Tac + CL  & 37.62 \\
    \midrule
    Patch Tac + CMA & 40.42 \\
    Patch Tac + GD  & 39.40 \\
    \textbf{GeoTAC} & \textbf{42.10} \\
    \bottomrule
    \end{tabular}
\end{center}
\end{minipage}
\end{figure}

Figure~\ref{fig:multi-sensory-result} presents the results using all multimodal sensory data. We observe that: (1) Incorporating more modalities requires stricter alignment, as all object-level results decline when additional modalities are involved. UniTouch's average score decreases by 9.55 compared with object-level alignment using CL. (2) Our proposed GeoTAC achieves the best results, improving the score by 4.48 over vision-tactile methods. (3) Correct labels are more important than complex alignment methods: Tac with contrastive learning (CL) and object-level multimodal alignment using ImageBind outperform other fine-grained methods. As seen in Figure~\ref{fig:attention}, geometry aligns the correct patch area, and other object-level methods show global importance and incorrect areas.

\subsection{Evaluation on Robotic Manipulation}
In this subsection, we explore the utility of tactile representations learned under different training strategies for robotic manipulation tasks. We replace the vision and tactile encoders with our trained models and post-train the policies on robot demonstrations.

Table~\ref{tab:univtac-policy-results} presents the results on robotic manipulation tasks. We observe the following: (1) The robotics results are consistent with the representation benchmark. Comparing Table~\ref{tab:univtac-policy-results} and Figure~\ref{fig:multi-sensory-result}, GeoTAC achieves the best results, UniTouch performs the worst, and the other methods show largely consistent rankings. (2) Poorly performing vision-tactile encoders yield worse results than vision-only encoders. UniTouch and ViT-Lens perform worse than Tac + CL in Figure~\ref{fig:multi-sensory-result}, and policies using their trained encoders perform worse than DINOv2 (Vision-only) in Table~\ref{tab:univtac-policy-results}. (3) Patch-aligned multisensory learning improves policy performance in simulation: Patch Tac + CL and GeoTAC improve average TP by 6.67 and 11.34 percentage points over DINOv2 (Vision-Tactile), respectively.

\begin{table}[!t]
\caption{Task performance in UniVTAC across multisensory learning algorithms.
Parentheses show changes in average SR and TP relative to
DINOv2 (Vision-only), in percentage points.}
\label{tab:univtac-policy-results}
\begin{center}
\setlength{\tabcolsep}{1.9pt}
\resizebox{\linewidth}{!}{%
\begin{tabular}{@{}l*{4}{cc}|cc@{}}
\toprule
\multirow{2}{*}{\makecell[l]{Encoder\\initialization}}
& \multicolumn{2}{c}{\makecell{Lift\\Can}}
& \multicolumn{2}{c}{\makecell{Lift\\Bottle}}
& \multicolumn{2}{c}{\makecell{Grasp\\Classify}}
& \multicolumn{2}{c}{\makecell{Insert\\Tube}}
& \multicolumn{2}{|c@{}}{Avg. (\%)} \\
\cmidrule(lr){2-3}
\cmidrule(lr){4-5}
\cmidrule(lr){6-7}
\cmidrule(lr){8-9}
\cmidrule(lr){10-11}
& SR & TP $\uparrow$
& SR & TP $\uparrow$
& SR & TP $\uparrow$
& SR & TP $\uparrow$
& SR $\uparrow$ & TP $\uparrow$ \\
\midrule
DINOv2 (Vision-only)
& 2/40 & 7.34
& 0/40 & 40.01
& 24/40 & 73.02
& 0/40 & 88.08
& 16.25 (0.00)
& 52.11 (0.00) \\

DINOv2 (Vision-Tactile)
& 4/40 & 9.32
& 2/40 & 50.45
& 24/40 & 90.00
& 0/40 & 89.90
& 18.75 \textcolor{green!50!black}{($+2.50$)}
& 59.92 \textcolor{green!50!black}{($+7.81$)} \\
\midrule
ImageBind~\citep{girdhar2023imagebind}
& 5/40 & 9.78
& 4/40 & 55.07
& 40/40 & 100.00
& 1/40 & 85.81
& 31.25 \textcolor{green!50!black}{($+15.00$)}
& 62.67 \textcolor{green!50!black}{($+10.55$)} \\

UniTouch~\citep{yang2024binding}
& 1/40 & 1.20
& 1/40 & 42.08
& 23/40 & 69.58
& 0/40 & 76.15
& 15.63 \textcolor{red}{($-0.63$)}
& 47.25 \textcolor{red}{($-4.86$)} \\

ViT-Lens~\citep{lei2024vitlens}
& 2/40 & 5.78
& 2/40 & 49.62
& 24/40 & 71.95
& 0/40 & 77.70
& 17.50 \textcolor{green!50!black}{($+1.25$)}
& 51.26 \textcolor{red}{($-0.85$)} \\
\midrule
Patch Tac + CL
& 6/40 & 10.08
& 4/40 & 65.51
& 40/40 & 100.00
& 8/40 & 90.77
& 36.25 \textcolor{green!50!black}{($+20.00$)}
& 66.59 \textcolor{green!50!black}{($+14.48$)} \\

\textbf{GeoTAC (Ours)}
& 6/40 & 20.01
& 5/40 & 72.38
& 40/40 & 100.00
& 12/40 & 92.63
& 39.38 \textcolor{green!50!black}{($+23.13$)}
& 71.26 \textcolor{green!50!black}{($+19.14$)} \\
\bottomrule
\end{tabular}%
}
\end{center}
\end{table}

\subsection{Ablation Study}

We conduct ablation studies on the utility of individual modules, specifically comparing performance when adding Cross-Modal Alignment (row \textbf{Patch Tac + CMA}) and Geometry Distillation (row \textbf{Patch Tac + GD}) separately. Table~\ref{tab:ablation-study} reports the average vision-tactile alignment score (the same as \textit{Avg.} in Table~\ref{tab:vision-tactile}). Building on patch alignment with contrastive learning (row \textbf{Patch Tac + CL}), adding CMA and GD improves the score by 2.8 and 1.78, respectively, with CMA providing a larger gain than GD. Combining both modules yields an improvement of 4.48 over the baseline.


\section{Limitations}

Despite our extensive efforts to collect 3D vision-tactile data and explore multisensory learning methods, several limitations remain: (1) limited aligned multisensory data; (2) challenges in unified learning across sensors; and (3) evaluation of utility to real-world robotic tasks.

\section{Conclusion}

We formalize tactile alignment with other modalities and build a 3D vision-tactile collection system, acquiring over 20K tactile samples from 505 objects. We extend existing tactile benchmarks from object- to patch-level alignment and propose GeoTAC to learn tactile representations from aligned multisensory data. The proposed methods exhibit consistent improvements in representation learning and robotic tasks.

\subsection*{AI use statement}

In this work, we used generative AI to correct grammar, polish sentences, and assist with figure creation.


\bibliography{iclr2027_conference}
\bibliographystyle{iclr2027_conference}

\appendix

\section{More Details on Touch3D}
\label{sec:supp-touch3d}

This section describes the annotation interface and object distribution of Touch3D, which contains 505 physical objects and 20,025 tactile contacts.

\subsection{Data Inspection and Annotation}

\begin{figure}[htbp]
\centering
\includegraphics[width=.88\linewidth]{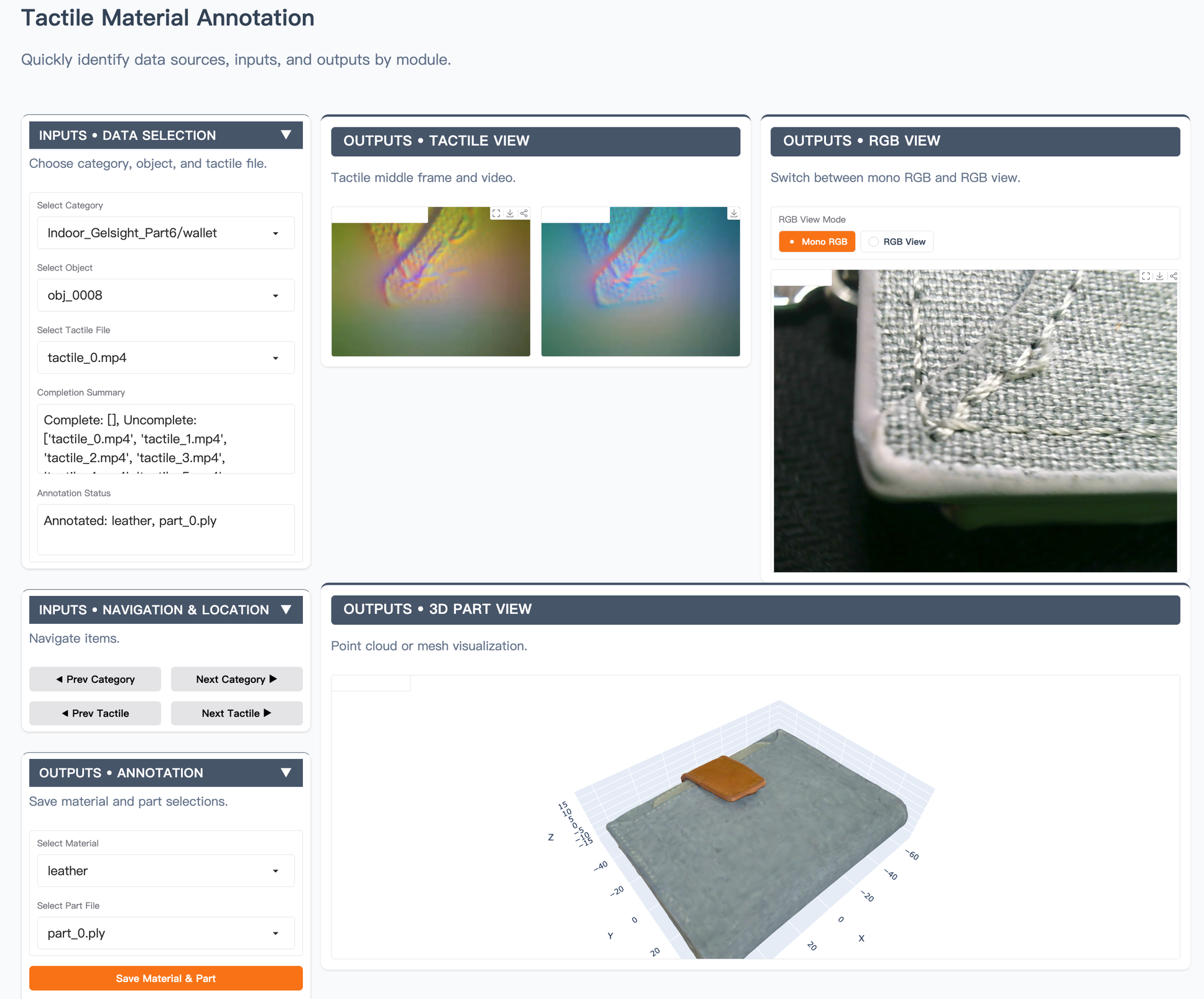}
\caption{\textbf{Data inspection and annotation interface.} Sample selection, navigation, and annotation controls are shown on the left; tactile, RGB, and 3D-part views support joint inspection of each contact.}
\label{fig:annotation}
\end{figure}

Figure~\ref{fig:annotation} shows the interface used to inspect collected samples and assign annotations. The left column selects the object category, instance, and tactile file, provides navigation between samples, and records material and 3D-part labels. The center and right panels display the tactile frame and video, the corresponding RGB view, and the scanned 3D part. Viewing these observations together helps operators check contact correspondence and material labels. As described in the main paper, incorrect geometry or contact annotations are corrected, while samples with missing or damaged observations are removed.

\subsection{Object Coverage and Distribution}

\begin{figure}[htbp]
\centering
\includegraphics[width=\linewidth]{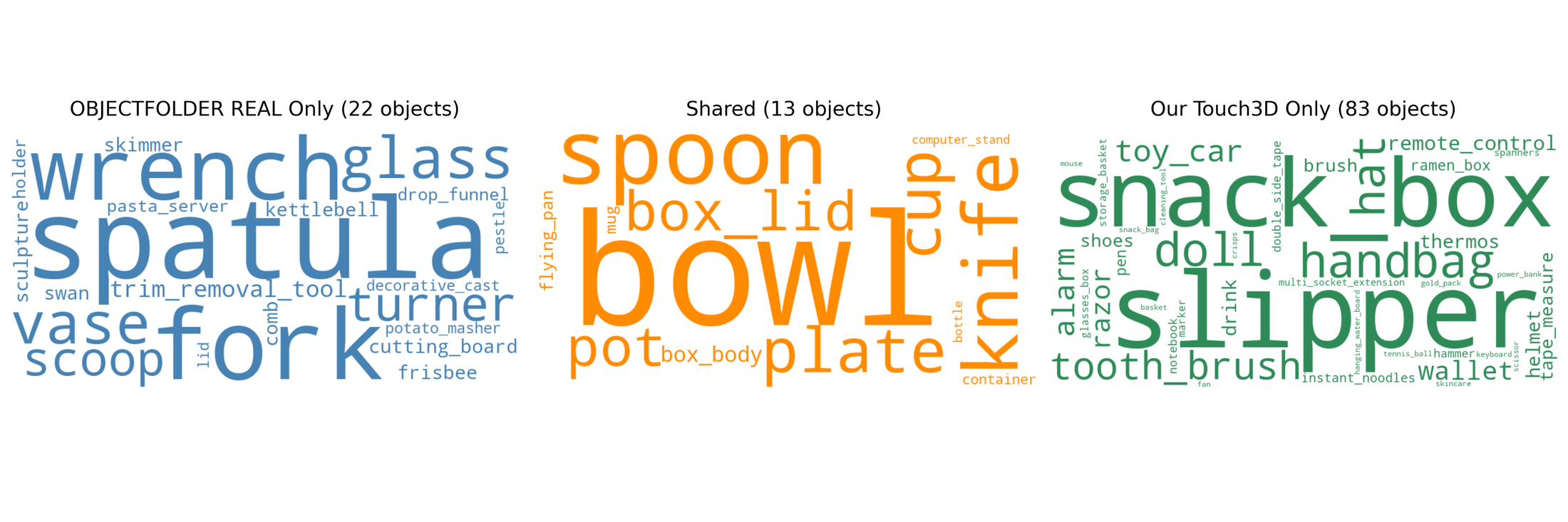}
\caption{\textbf{Object Category Coverage.} From left to right: names unique to ObjectFolder Real, shared by both datasets, and unique to Touch3D. The panel counts describe distinct names rather than physical object instances.}
\label{fig:word-cloud}
\end{figure}

Figure~\ref{fig:word-cloud} compares the object names represented in Touch3D and ObjectFolder Real. The three groups contain 22 categories unique to ObjectFolder Real, 13 shared categories, and 83 categories unique to Touch3D. Shared categories include bowls, spoons, cups, and plates, while Touch3D additionally covers objects such as slippers, handbags, toy cars, and toothbrushes. These counts refer to distinct object names in the comparison, not individual physical instances.

\begin{figure}[htbp]
\centering
\includegraphics[width=\linewidth]{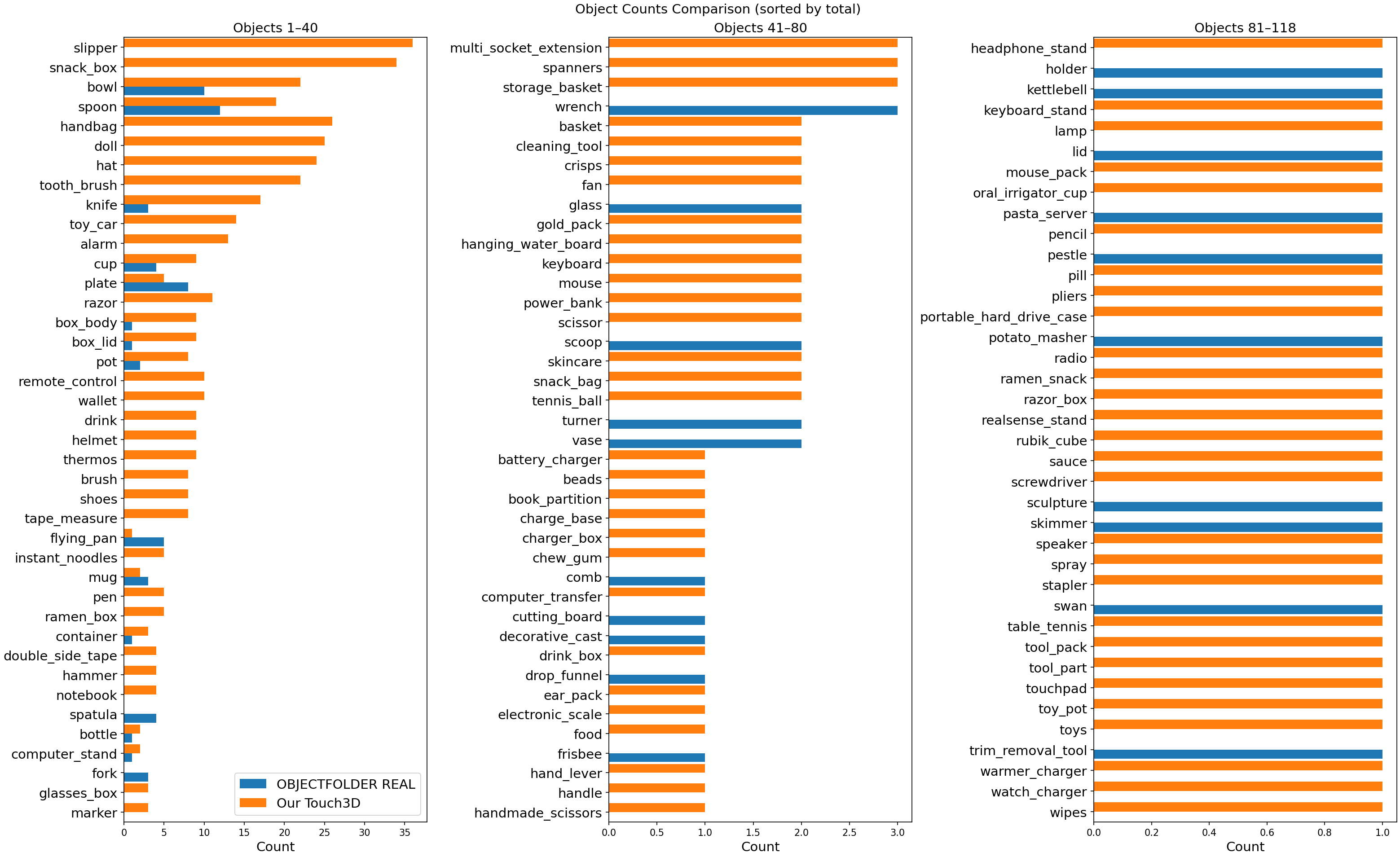}
\caption{\textbf{Object-instance counts in Touch3D and ObjectFolder Real.} Categories are sorted by combined count across both datasets. The three panels use different horizontal scales to show frequent and infrequent categories.}
\label{fig:object-count}
\end{figure}

Figure~\ref{fig:object-count} complements this comparison with instance counts for each object category. Blue and orange bars represent ObjectFolder Real and Touch3D, respectively. Categories are ordered by their combined counts and divided into three panels with different horizontal scales. Both datasets have uneven category frequencies: a few categories account for many instances, while numerous others have only one or a few. Thus, the balanced coverage of the six collection scenarios described in the main paper does not imply balanced counts for every object category.

\section{Additional Evaluation Details}
\label{sec:supp-evaluation}

\subsection{Representation Learning and Evaluation}

We use DINOv2 ViT-Large for the patch-alignment experiments. Touch3D is split into training, validation, and test sets at a ratio of 4:1:1, and ObjectFolder Real provides an additional test set. Representation training uses Touch3D; evaluation on ObjectFolder Real examines transfer to data collected with a different tactile sensor.


Object-level evaluation measures category correspondence between a local tactile query and global visual candidates using k-NN Top-5 accuracy. Patch-level evaluation instead requires the candidate to originate from the same contact and reports exact-contact Top-5 retrieval accuracy (E@5), averaged over both directions. Matching the object category alone is therefore insufficient for patch-level success. Material prediction uses k-NN mAP with multiple material labels, while object-category classification uses probe accuracy. The average in Table~\ref{tab:vision-tactile} and Table~\ref{tab:ablation-study} is the unweighted mean of these four vision-tactile evaluation metrics.

\subsection{Robotic Manipulation Evaluation}

We evaluate transfer to Diffusion Policy in the UniVTAC simulator using Lift Can, Lift Bottle, Grasp Classify, and Insert Tube. Each task uses 100 demonstration episodes and 30,000 policy-training steps, followed by 40 evaluation rollouts. The vision-only baseline excludes tactile observations; the vision-tactile methods use both modalities with their corresponding encoder initializations.

In Table~\ref{tab:univtac-policy-results}, task-level SR is the number of successful rollouts out of 40. TP is the mean final task progress across all 40 rollouts, including unsuccessful ones. Average SR and TP weight the four tasks equally and are computed before rounding. Parenthesized changes are absolute percentage-point differences from DINOv2 (Vision-only), not relative percentage gains. These results measure simulated manipulation rather than physical-robot deployment.

\section{Limitations and Future Directions}
\label{sec:supp-limitations}

\paragraph{Data Scale and Alignment Quality.}
Collecting paired tactile observations and scanned geometry remains costly, and category coverage is uneven. Generated geometry~\citep{wang2025vggt,xiang2025structured}, relabeled object-level data, and tactile simulation~\citep{nguyen2024tacex} could expand the training set, but their contact correspondences require quality control before being used as patch-level supervision.

\paragraph{Cross-Sensor Generalization.}
Differences in sensor appearance, geometry, and response can limit transfer between datasets. Learning from multiple tactile sensors and harmonizing contact annotations are important directions toward a shared tactile representation.

\paragraph{Physical-Robot Transfer.}
The current manipulation evaluation is conducted in simulation. Physical-robot experiments are still needed to assess robustness to real sensor noise, calibration errors, and variations in contact. They would also help determine when improved representation metrics translate into more reliable manipulation.

\end{document}